\documentclass[10pt,letterpaper]{article}

\usepackage{amsmath,amssymb,subfig}

\usepackage{changepage}

\usepackage[utf8x]{inputenc}

\usepackage{textcomp,marvosym}

\usepackage{cite}

\usepackage{nameref,hyperref}

\usepackage[table]{xcolor}

\usepackage{array}

\newcolumntype{+}{!{\vrule width 2pt}}

\newlength\savedwidth

\usepackage{tikz}

\usetikzlibrary{calc,trees,positioning,arrows,chains,shapes.geometric,decorations.pathreplacing,decorations.pathmorphing,shapes,    matrix,shapes.symbols}

\tikzset{
>=stealth',
  punktchain/.style={
    rectangle, 
    rounded corners, 
    draw=black, very thick,
    text width=25em, 
    minimum height=3em,
    minimum width = 20em,
    text centered, 
    on chain},
  line/.style={draw, thick, <-},
  element/.style={
    tape,
    top color=white,
    bottom color=blue!50!black!60!,
    minimum width=8em,
    draw=blue!40!black!90, very thick,
    text width=10em, 
    minimum height=3.5em, 
    text centered, 
    on chain},
  every join/.style={->, thick,shorten >=1pt},
  decoration={brace},
  tuborg/.style={decorate},
  tubnode/.style={midway, right=2pt},
}

\raggedright
\makeatletter
\renewcommand{\@biblabel}[1]{\quad#1.}
\makeatother

\date{}

\newtheorem{thm}{Theorem}

\begin{document}
\vspace*{0.2in}

\begin{flushleft}
{\Large
\textbf\newline{Calculating the Midsagittal Plane for Symmetrical Bilateral Shapes: Applications to Clinical Facial Surgical Planning } 
}
\newline
\\
Aarti Jajoo\textsuperscript{1\Yinyang *},
Matthew Nicol \textsuperscript{2\Yinyang},
Jaime Gateno\textsuperscript{3\Yinyang},
Ken-Chung Chen\textsuperscript{3\Yinyang},
Zhen Tang\textsuperscript{3\Yinyang},
Tasadduk Chowdhury\textsuperscript{2\Yinyang},
Jainfu Li\textsuperscript{2\Yinyang},
Steve Goufang Shen\textsuperscript{4\Yinyang},
James J. Xia\textsuperscript{3\Yinyang}
\\
\bigskip
\textbf{1}  Molecular and Human Genetics, Baylor College of Medicine, Houston, Texas, United States
\\
\textbf{2}  Department of Mathematics, University of Houston, Houston, Texas, United States
\\
\textbf{3} Oral and Maxillofacial Surgery Department, Houston Methodist Hospital, Houston, Texas, United States
\\
\textbf{4} Department of Oral and Craniomaxillofacial Surgery, Shanghai Ninth People's Hospital, Shanghai, China
\bigskip

%
%
\Yinyang These authors contributed equally to this work.





* jajoo@bcm.edu

\end{flushleft}
\section*{Abstract}
It is difficult to estimate the midsagittal plane of human subjects with craniomaxillofacial (CMF) deformities. We have developed a LAndmark GEometric Routine (LAGER), which automatically estimates a midsagittal plane for such subjects. The LAGER algorithm was based on the assumption that the optimal midsagittal plane of a patient with a deformity is the premorbid midsagittal plane of the patient (i.e. hypothetically normal without deformity). The LAGER algorithm consists of three steps. The first step quantifies the asymmetry of the landmarks using a Euclidean distance matrix analysis and ranks the landmarks according to their degree of asymmetry. The second step uses a recursive algorithm to drop outlier landmarks. The third step inputs the remaining landmarks into an optimization algorithm to determine an optimal midsaggital plane. We validate LAGER on 20 synthetic models mimicking the skulls of real patients with CMF deformities. The results indicated that all the LAGER algorithm-generated midsagittal planes met clinical criteria.  Thus it can be used clinically to determine the midsagittal plane for patients with CMF deformities.

\section*{Introduction}
\label{intro}
Three-dimensional (3D) computed tomography (CT) models are used not only  for diagnosing  craniomaxillofacial (CMF) deformities, but also for planning  surgeries. During the planning process, it is crucial to position the 3D CT models in a unique reference frame. Routinely, the standard anatomical axial, coronal and midsagittal planes are well suited to this purpose. These planes are orthogonal and are aligned to the world axes of Euclidean space. The standardized use of a reference frame requires that the object and the frame are aligned with each other. Defining the axial and coronal planes can be done by aligning the face to the reference frame, while defining the midsagittal plane divides the face into right and left halves. 

To improve the accuracy of the anatomical plane determination, our medical collaborators have used a method of recording head orientation, i.e. natural head position (NHP), using digital orientation sensors~\cite{XMG11,SXG10}. The 3D CT model was reoriented using the recorded NHP as the world frame of the reference. The 3D CT model is translated to the world frame of the reference with the soft tissue nasion landmark passing through the midsagittal plane, which divides the face into the right and left parts. However, human head posture has variability. Within subject repeated measurements of head orientation can vary as much as $\pm$ 2 degrees~\cite{MST08,MOO58,LUN92}. Even when the repeated measurements are averaged, this variation still causes clinical problems when determining the midsagittal plane, as a 2 degree offset in roll (the rotation around the anteroposterior axis) may result in iatrogenic transverse misalignment of a facial unit by as much as 4mm. Hence, we started exploring other potential methods that can be used to determine the midsagittal plane. 

Theoretically, the midsagittal plane can be simply determined by the use of anatomical midline landmarks, which are ideally located on the mid-sagittal plane of a normal subject. However, a third of our patients have gross facial asymmetry that warps this plane. We initially assumed that a solution would be to create a best-fit plane among all these midline landmarks. Unfortunately, this approach did not work. This is best explained using an example. Imagine we have a patient with unilateral condylar hyperplasia in that one side of the mandible and condyle is larger  than the contralateral side. Such patients were born normal and have normal form and symmetry until adolescence when one of the mandibular condyles grows asymmetrically, producing severe deviation of the chin and dental midline. The goal of surgery is to return this patient to his premorbid condition. For this we would have to use his premorbid midsagittal plane as our reference. Finding a plane of best fit for the midline landmarks may produce a different plane, resulting in a less than ideal surgical outcome. Therefore, the purpose of this study was to develop a mathematical algorithm that can automatically determine the midsagittal plane for a patient, the midsagittal plane he would have if his face was not deformed. 

\section*{Landmark Geometric Routine (LAGER) }\label{sec:lager}
\begin{figure}[!t]
\centering
\subfloat[Digitized Landmarks]{\includegraphics[width=2.5in]{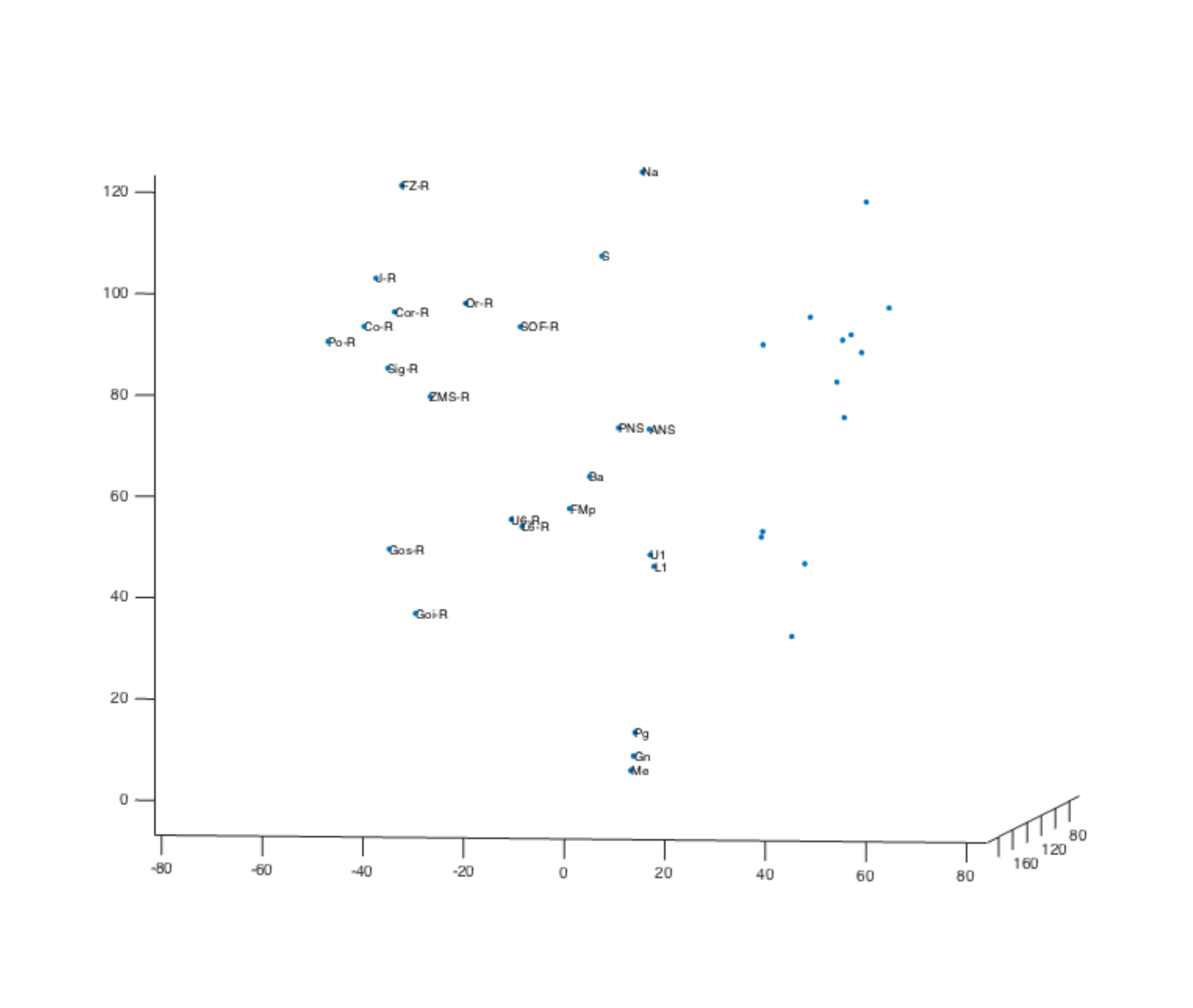}}
\hfill
\subfloat[Landmarks and Skull]{\includegraphics[width=2.5in]{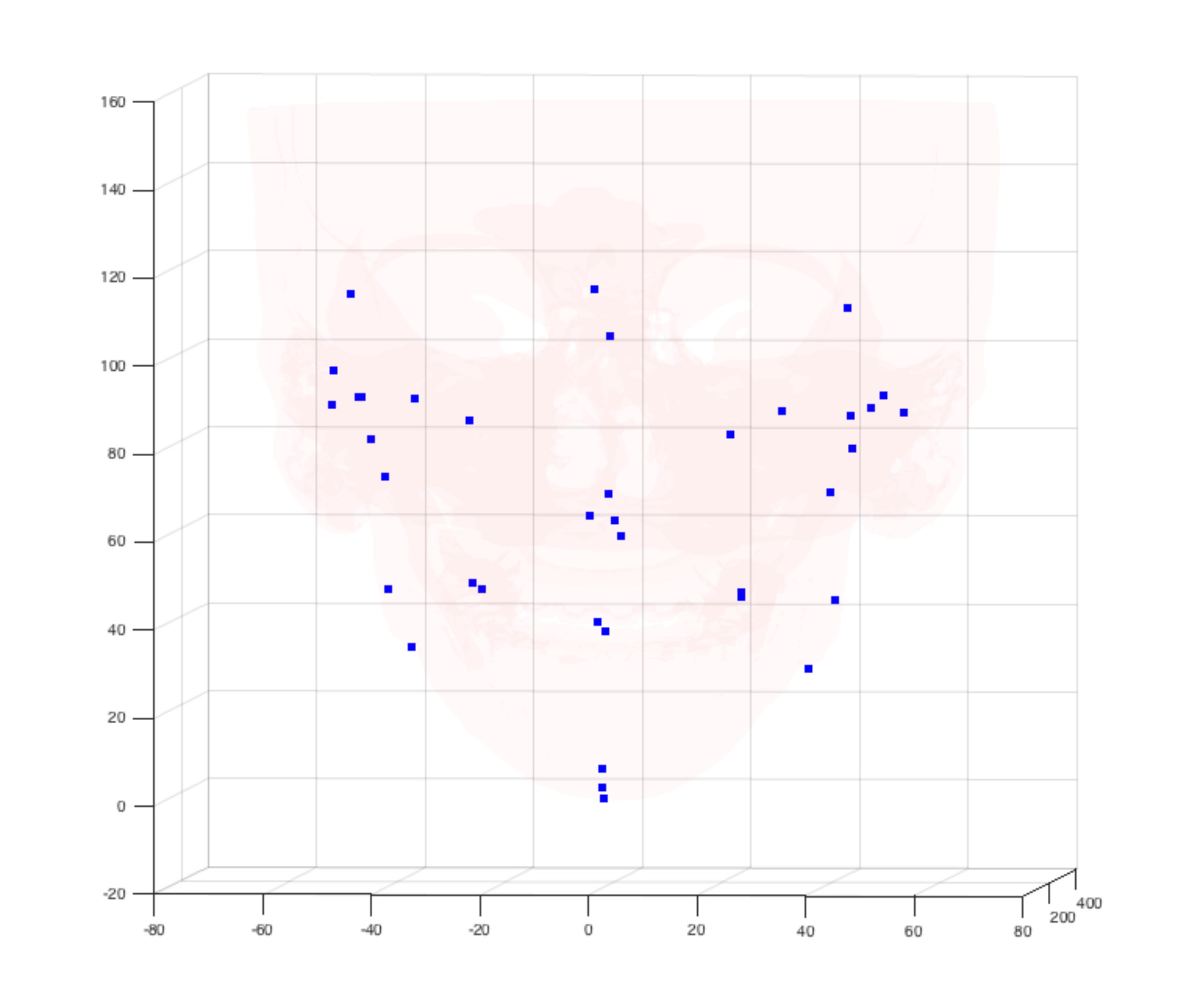}}%
\caption{Anatomical landmarks digitized for a patient. Left figure: 2D view of landmarks; Right figure: landmarks and superimposed 3D CT model.}
\label{fig:Landmark}
\end{figure}

We have developed a {\em LAndmark GEometric Routine} ({\bf LAGER}) to determine the midsagittal plane for a deformed face. LAGER takes cephalometric landmarks as input. A list of standard cephalometric landmarks can be found here~\cite{SSH06}. Broadly speaking there are two types of landmarks: unpaired and paired. In the ideal case unpaired landmarks are located on the midsagittal plane while the paired ones are located on the right and left sides of the face in pairs symmetric with respect to the midsagittal plane.  For the scope of this paper, we use 11 unpaired and 26-paired landmarks (corresponding to 13 pairs), see Table~\ref{tb:landmarkpoints} and Fig \ref{fig:Landmark}. However, our algorithm does not depend on the choice of landmarks and any other landmark set that represent important cephalometric features of a face can be used~\cite{SSH06}. 

\begin{figure}[!t]
\centering
\subfloat[Perfect symmetry]{\includegraphics[width=1.5in,trim={4.5cm 3cm 4.5cm 3cm},clip]{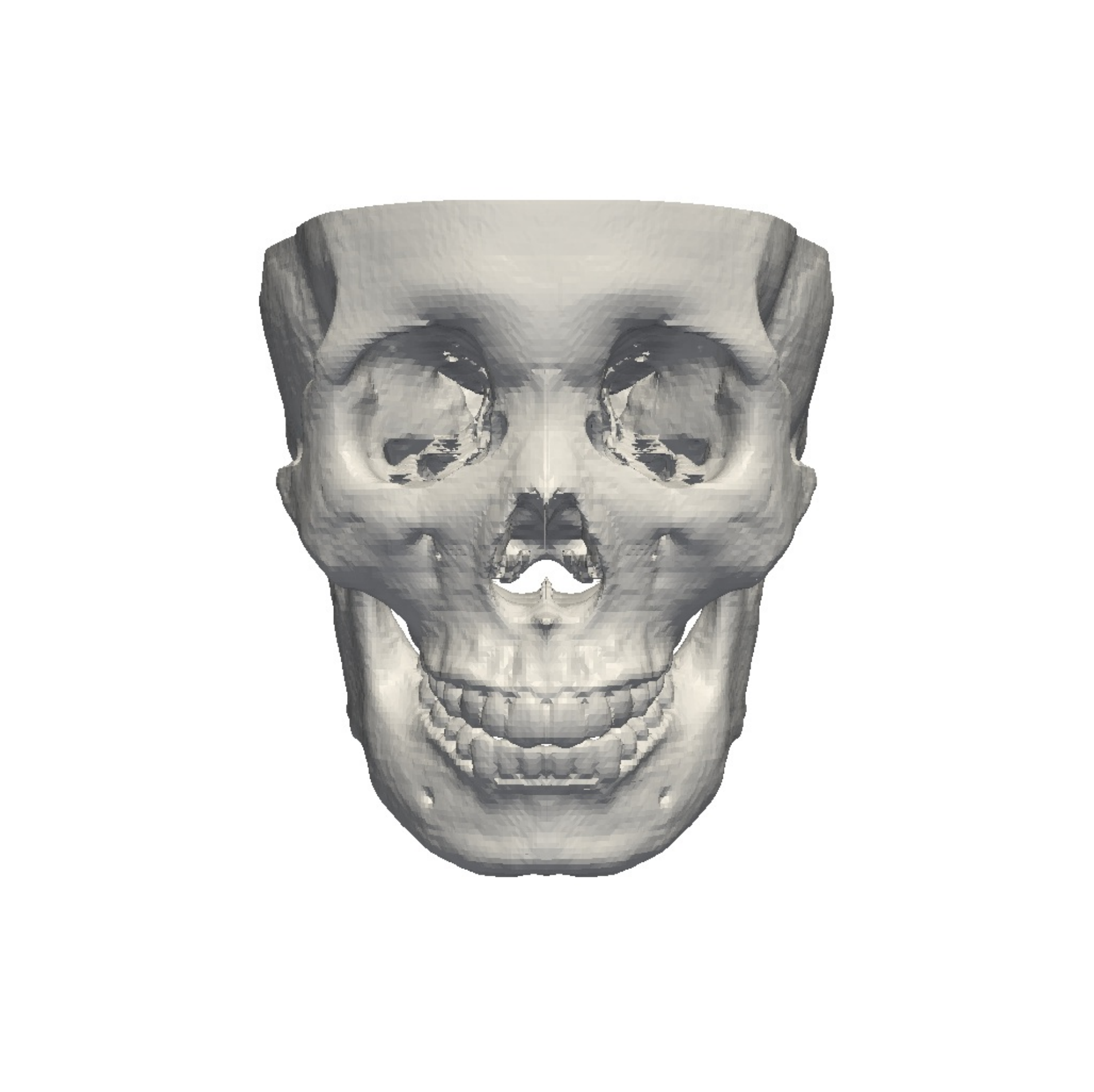}
\label{fig:normalcasesymmetric}}
\hfill
\subfloat[Fluctuating asymmetry]{\includegraphics[width=1.4in,trim={4.5cm 3cm 4.5cm 3cm},clip]{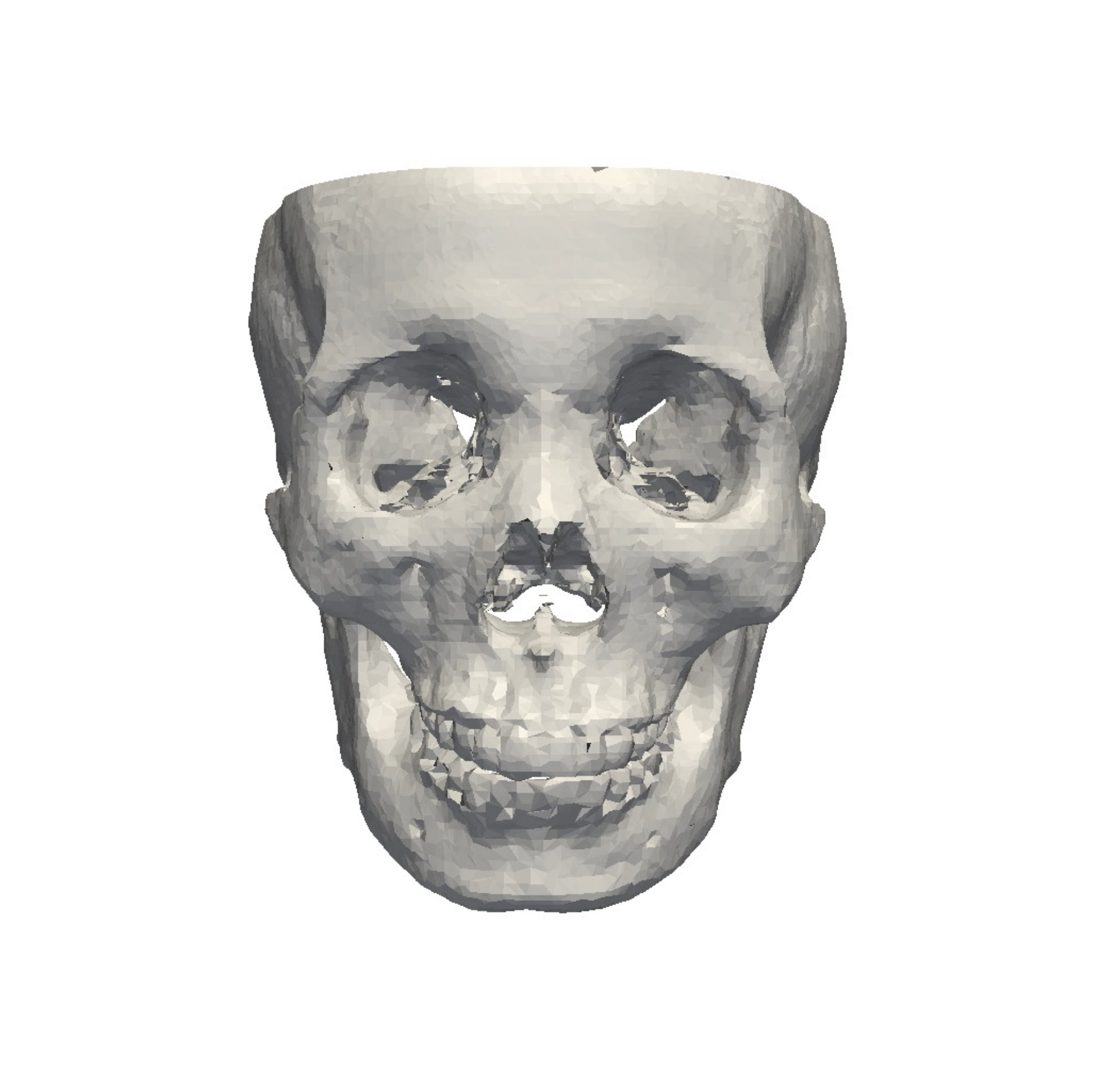}%
\label{fig:normalcase}}
\hfill
\subfloat[Deformity asymmetry]{\includegraphics[width=1.5in,trim={4.5cm 3cm 4.5cm 3cm},clip]{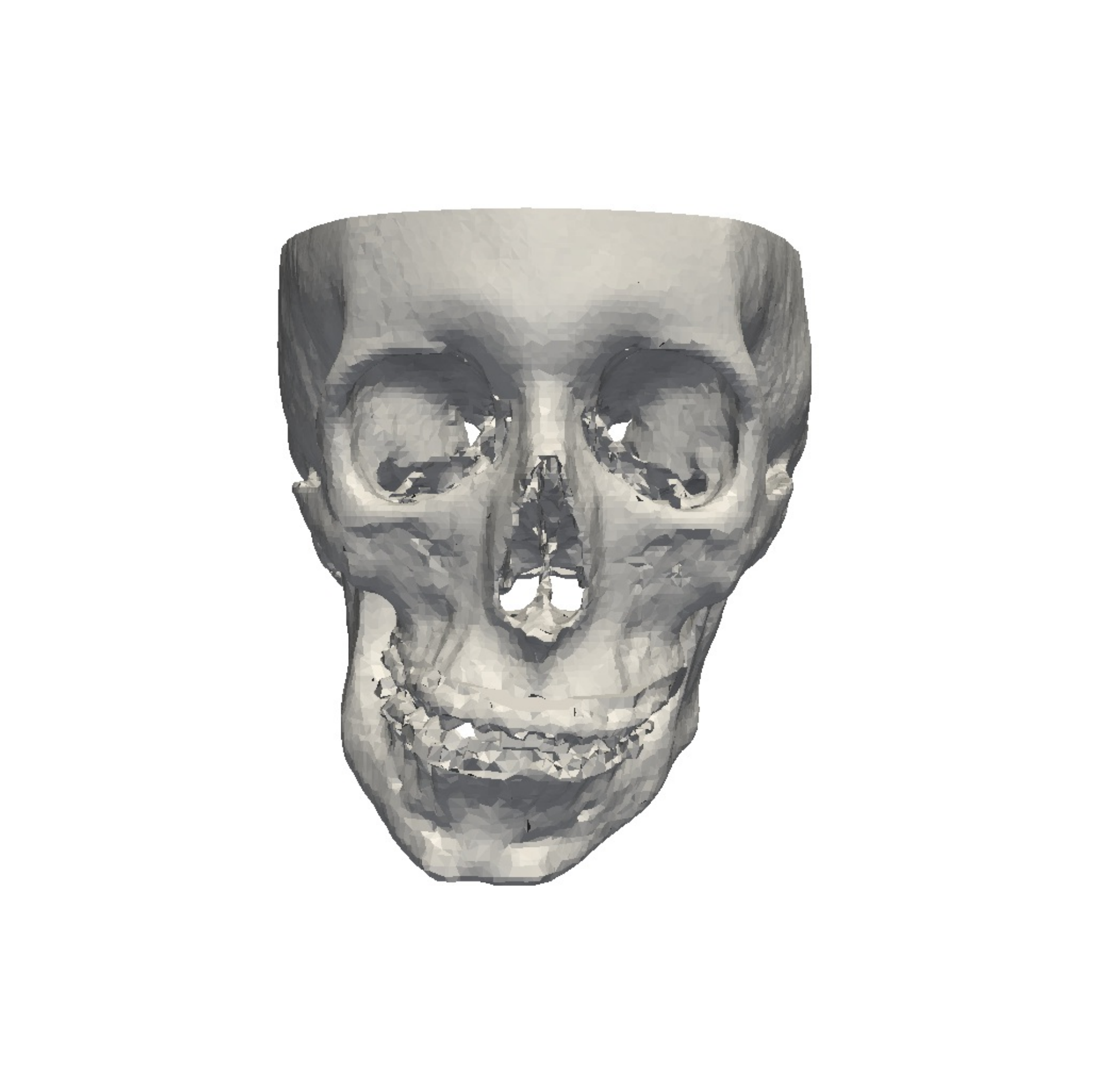}%
\label{fig:deformedcase}}
\caption{Skull models obtained by the facial 3D CT models:  The skull in the middle (b)  is of a normal person, notice that  the skull is approximately symmetric but not completely symmetric. The midsagittal plane for this skull is apparent and divides the skull in almost identical right and left halves. The skull to the left (a) is a perfectly symmetric skull generated by replacing the left side of the normal skull by its right side. Notice that the right side of both skulls is the same. The figure to the right (c) is the skull of a patient with type-2 hemifacial microsomia. For this patient the choice of a  midsagittal plane is not obvious.}
\label{fig:CTtoBone}
\end{figure}

For this study, 2 oral surgeons manually placed landmarks on 3D skull models derived from the facial CT models of a patient. Fig~\ref{fig:normalcase} shows the skull of a normal subject. Studies show that even for a normal subject, there always exist minor asymmetries of the face, known as fluctuating asymmetry~\cite{LEI62}. Yet, the midsagittal plane for a normal subject is fairly apparent. However, the same is not true for patients with congenital or acquired CMF deformities, e.g. a patient with hemifacial microsomia, Fig~\ref{fig:deformedcase}.  Hence, we need an algorithm, which can automatically calculate the midsagittal plane for such patients. 

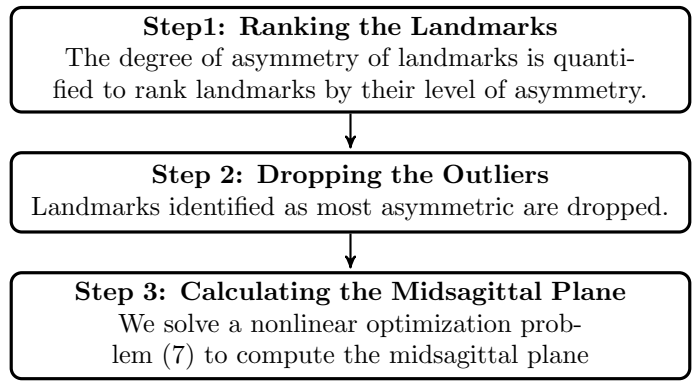
\begin{figure} [h]
\centering
\begin{tikzpicture}
  [node distance=.5cm,
  start chain=going below,]
     \node[punktchain, join] (probf)      {{\bf Step1: Ranking the Landmarks} \\ The  degree of asymmetry of landmarks is quantified to rank landmarks by their level of asymmetry.};
     \node[punktchain, join] (probf) {{\bf Step 2: Dropping the Outliers} \\ Landmarks identified as most asymmetric are dropped.};
     \node[punktchain, join] (investeringer){ {\bf Step 3: Calculating the Midsagittal Plane}\\ We solve a nonlinear optimization problem (\ref{equ:objfun}) to compute the midsagittal plane};
        \end{tikzpicture}
       \caption{LAGER Algorithm}
       \label{flw:flowchart}
       \end{figure}

Our algorithm leverages the fact that most facial deformities do not affect the whole face of a patient. One could use the landmarks that represent unaffected or least affected regions of a face to determine the plane. But, due to the presence of fluctuating asymmetry, it is not apparent how to identify such regions. To this end, we have developed LAGER algorithm, which involves three steps, Fig~\ref{flw:flowchart}. In the first step, we quantify asymmetry level of each landmark and rank them by their estimated level of asymmetry. In the second step, based on this quantification of the asymmetry of landmarks, we classify a set of landmarks as outliers. In other words, we identify landmarks that represent regions, which are most affected by the deformity. We then drop this set of landmarks from the original set of landmarks. In the third step, we calculate the midsaggital plane using remaining landmarks. We explain these steps in detail in the following subsections.  
\begin{table*}[!ht]
\begin{adjustwidth}{-2.25in}{0in} 
\caption{Landmark Points Used by LAGER Algorithm to Calculate the Midsagittal Plane} \label{tb:landmarkpoints}
\centering
\begin{tabular}{c |c |l }
\hline
&\textbf{Unpaired Landmarks (midline)}		& Anatomical definition\\ \hline			
1	&	Sella (S)	&	The midpoint of the  sella turcica on the midsagittal plane	\\ \hline
2	&	Nasion (N)	&	The most anterior point on fronto-nasal suture	\\ \hline
3	&	Basion (Ba)	&	The most anterior point of foramen magnum	\\ \hline
4	&	FMp	&	The most posterior point of foramen magnum	\\ \hline
5	&	Anterior Nasal Spine (ANS)	&	The most anterior midpoint of the anterior nasal spine of the maxilla	\\ \hline
6	&	Posterior Nasal Spine (PNS)	&	The most posterior midpoint of the posterior nasal spine of the palatine bone	\\ \hline
7	&	Upper Central Dental Midline (U1)	&	The midpoint between the right and left upper central incisal edges	\\ \hline
8	&	Lower Central Dental Midline (L1)	&	The midpoint between the right and left lower central incisal edges	\\ \hline
9	&	Pogonion (Pg)	&	The most anterior midpoint of the chin on the mandibular symphysis	\\ \hline
10	&	Gnathion (Gn)	&	The most anteroinferior midpoint of the chin on the mandibular symphysis	\\ \hline
11	&	Menton (Me)	&	The most inferior midpoint of the chin on the mandibular symphysis	\\ \hline			& \textbf{Paired Landmarks (right and left)}				& 	\\ \hline
 \hline
12	&	Orbitale (Or)	&	The most inferior point of the orbital rim	\\ \hline
13	&	Frontozygomatic Point (Fz)	&	The most medial and anterior point of the frontozygomatic suture at the level of  \\ & & the lateral orbital rim	\\ \hline
14	&	Jugale (Point J)	&	The intersection point formed by the masseteric and maxillary edges of the \\
& &  zygomatic bone	\\ \hline
15	&	Supraorbital Fissue (SOF)	&	The most superior and medial point of superior orbital fissue	\\ \hline
16	&	Porion (Po)	&	The most superior point on the bony external acoustic meatus	\\ \hline
17	&	zygomaticomaxillary suture (ZMS)	&	The most inferior and lateral point of zygomaticomaxillary suture	\\ \hline
18	&	U6	&	The mesiobuccal cusp of the upper first molar	\\ \hline
19	&	L6	&	The mesiobuccal cusp of the lower first molar	\\ \hline
20	&	Condylion (Co):	&	The most posterosuperior point of the mandibular condyle on the lateral view	\\ \hline
21	&	Sigmoid Notch (SIG):	&	The deepest point on the  mandibular sigmoid notch	\\ \hline
22	&	Coronion (Cr)	&	The tip of the coronoid process	\\ \hline
23	&	Gos	&	The most superoposterior point at the mandibular gonial angle	\\ \hline
24	&	Goi	&	The most inferior point at the mandibular gonial angle	\\ \hline
\end{tabular}
\end{adjustwidth}
\end{table*}

\subsection*{Step 1: Ranking the landmarks} \label{sec:rankinglandmark}
We rank the landmarks in order of asymmetry.  We use the Euclidean distance matrix analysis (EDMA) to quantify asymmetry as explained in~\cite{RIC96}.  
We denote coordinates of unpaired landmarks as $U_i, i=1\dots M$, the coordinates of paired landmarks of the right side as $P_i^R,i=1\dots N$ and the associated coordinates of landmarks of the left side as $P_i^L,i=1\dots N$, where $M$ is the number of unpaired landmarks and $N$ is the number of paired landmarks. 
We now consider two non-disjoint sets of landmarks: $L$ and $R$. $L$ contains paired landmarks of the left side and all the unpaired landmarks, $R$ contains paired landmarks of the right side and all the unpaired landmarks. 
We then compute the distances as defined below:
\begin{eqnarray}
(M^U_R)_{ij} 
&:=  &
\|U_i-P^R_j\|, \;\;i  =1\dots M, \;\;j=1 \dots N \\
(M^P_R)_{ij}
&:= &
\|P^R_{i}-P^R_j\|,\;\;1 \le i <j \le N 
\end{eqnarray}
Analogous matrices $M^U_L$ and $M^P_L$ associated to the left side are computed in the same fashion. $M^P_L$ and $M^P_R$ are symmetric matrices with diagonal elements zero, so it suffices to consider the elements above the diagonal or below. We define two vectors $V_R$ and $V_L$ as follows
\begin{multline}
V_R = \{ \textrm{All the entries of } M^U_R, \;\; \\ \textrm{All the upper half off diagonal entries of } M^P_R \}
\end{multline}
(Remark: $V_R$ is a vector of length $\frac{N(N-1)}{2}+MN$. )
An analogous vector $V_L$ associated to the left side is also constructed, such that the corresponding order of the entries is the same for both vectors, $V_R$ and $V_L$. We now compute the following ratio vector $(H)$:
\begin{equation}\label{equ: R}
(H)_{i}: = \left \|  \log \left ( \frac{(V_R)_i}{(V_L)_i} \right ) \right  \| 
\end{equation} 
Unlike other EDMA procedures, we use logarithm of ratios of elements of $V_R$ and $V_L$ because we are interested in capturing side invariant differences between the scores.  For example if one side score is 1 and another side is .8, then regardless of which side is left or right $H_i$ would always be $\| \log(\frac{1}{.8})\|$, as oppose to just using ratios. In which case the scores would be $\frac{1}{.8}$ and $\frac{.8}{1}$, when right side is $1$ and $.8 $ correspondingly. 

Clearly, if a model is perfectly symmetric then all the entries of the vector $H$ are zero. However, in a real clinical situation, we would have asymmetric models with landmarks having different levels of asymmetry, even in normal subjects.  It is reasonable to expect that the entries associated with a relatively more asymmetric landmark would be larger (in a relevant norm) than the norm of the entries associated with a relatively symmetric landmark. In order to see the effect of a landmark on the mean of the ratio vector $H$, we drop that landmark and recompute the mean of the resulting vector $H$ for the remaining set of landmarks. Note that when we drop a midline landmark we eliminate $N$ entries in $(H)$ and when we drop a paired landmark we eliminate $M+N$ entries in  $H$. More precisely, 
\begin{multline}
H_{(-i)}:= \{ \textrm{vector $H$ computed using all landmarks but for  } \\  
\textrm{ the ith landmark that we drop} \} \\
T_{(-i)} :=\{ \textrm{ mean of vector $H_{(-i)}$ } \},\; i=1,2, \; \dots, \; M+N 
\label{equ:Ti}
\end{multline}
A low value of $T_{(-i)}$ indicates that removing the associated landmark  decreases the asymmetry of the sets of points (as measured by the mean of the vector $H$). Hence it is reasonable to consider the landmark associated with the lowest $T_{(-i)},\; i = 1\dots,\; M+N$ to be the most asymmetric landmark. We now remove this most asymmetric landmark from the set of landmarks and consider the remaining set of landmarks as a new system. We then apply the same steps explained above to compute $T_{(-i)}$ scores for this new system to get the next most asymmetric landmark. We continue this recursive process until we have ranked all the landmarks. Notice that if the most asymmetric landmark turns out to be a paired landmark, then we remove the pair from the set of landmarks. 

$T_{(-i)}$ can also be used to classify severity of deformity, see Fig \ref{fig:HDistribution}. For normal subjects $T_{(-i)}$ usually ranges between .01 to .02. For mild and moderate deformities, such as mild Horizontal (Vertical) Condylar Hyperplasia or type-1 Hemifacial Microsomia $T_{(-i)}$ ranges between .01 to .04. For sever deformities (type 2 hemifacial microsomia) $T_{(-i)}$ ranges between .04 to .2. 

\subsection*{Step 2: Dropping the outliers} \label{sec:outlierdetection}
\begin{figure}[!t]
\centering
\subfloat[$T_{(-i)}$ score distribution]{\includegraphics[width=2.5in]{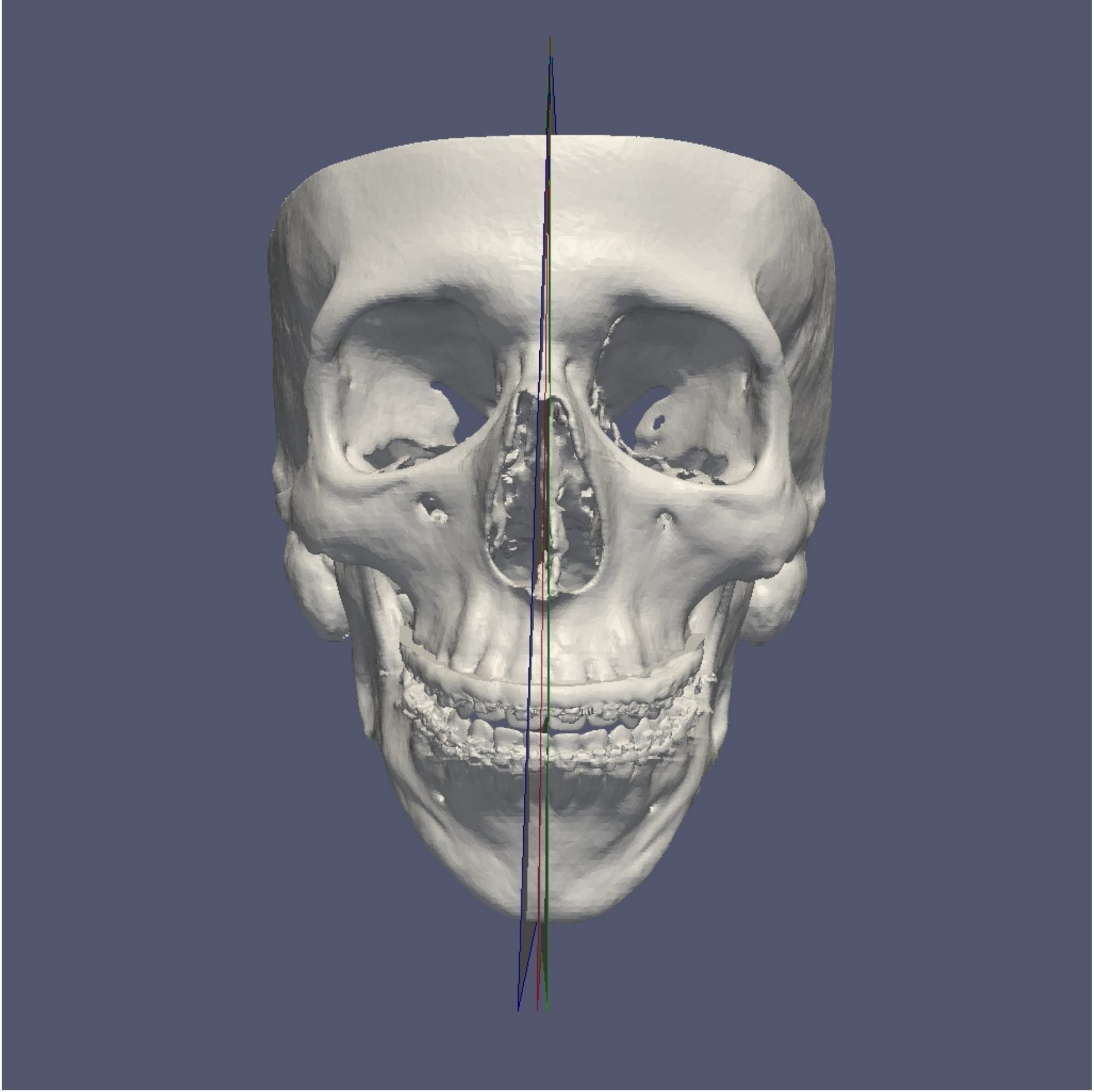}}
\hfill
\subfloat[Midsagittal planes]{\includegraphics[width=3in]{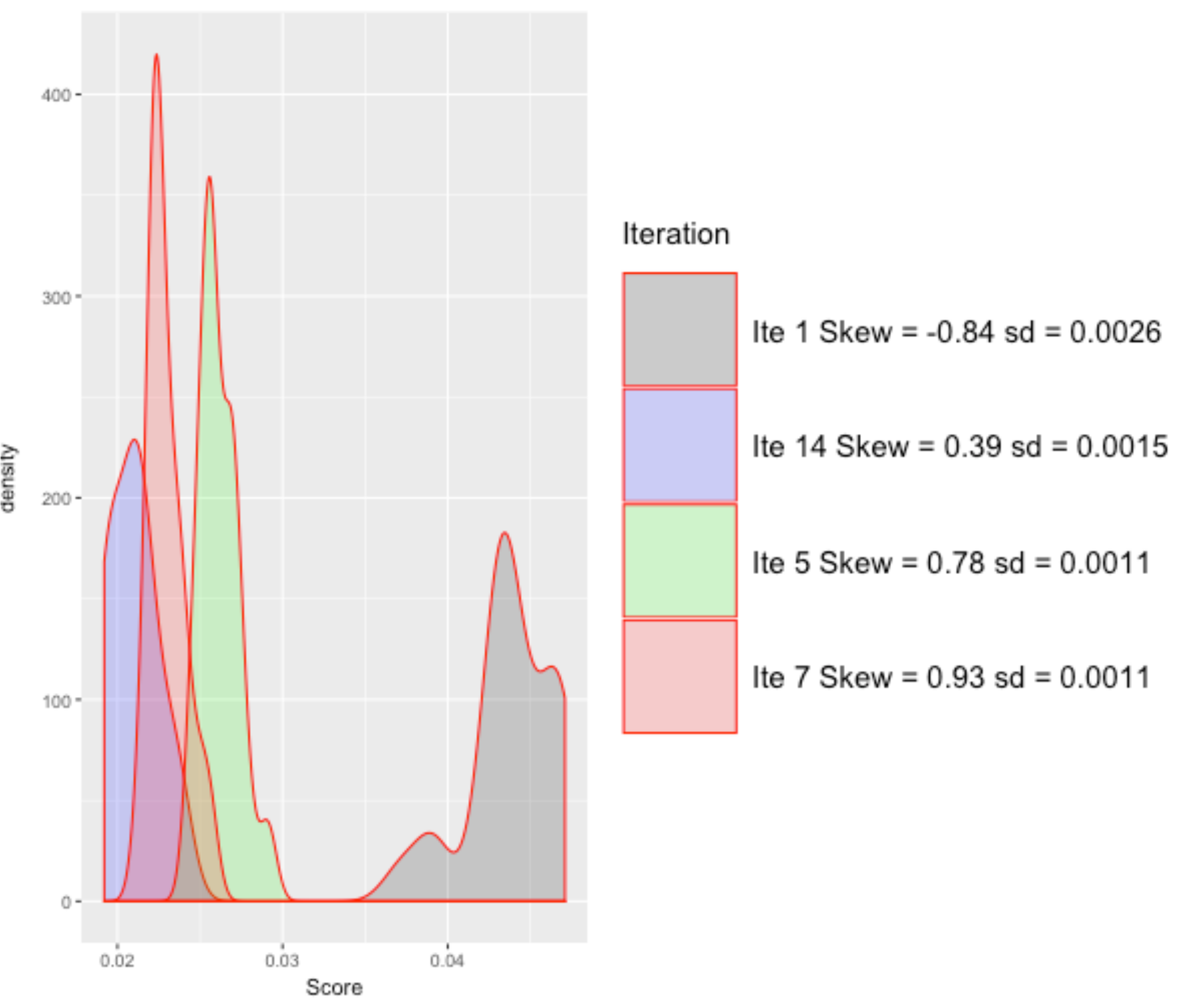}} 
\caption{In this example all the facial regions are deformed and hence several landmarks are affected. Left figure shows distribution of $T_{(-i)}$  at different iterations. Higher score at iteration 1 indicates that it is a case of sever asymmetry. As we keep dropping outliers, the scores start to decrease but not the standard deviation (sd) and skewness. In the right figure Blue plane is obtained at the 14th iteration, i.e. after dropping 14 outliers. But it gets biased to the remaining landmarks. Analysis of variance and skewness depicts that after dropping 5 outliers, the remaining landmarks are pretty much of the same order of asymmetry, see Step 2. Hence, the algorithm picks Green plane as the midsagittal plane.}
\label{fig:Patient29} 
\end{figure}

Outliers are determined sequentially in the order of landmark ranking that was calculated in Step 1, starting with identifying the most asymmetric landmark as the first outlier, then second most asymmetric landmark as the second outlier and so on. We continue this process till we identify all the outliers. Since none of the parts on the skull of a real patient are perfectly symmetric, $T_{(-i)}$ scores will never be perfect. This pose a problem, namely, how to know when to stop classifying landmarks as outliers. This is specially crucial when most of the facial regions are affected with deformity, in which case falsely identified outliers can bias the midsagital plane calculation towards the rest of the landmarks present in the system, see Fig \ref{fig:Patient29}.  

\begin{figure}[!t]
\centering
\includegraphics[width=5in]{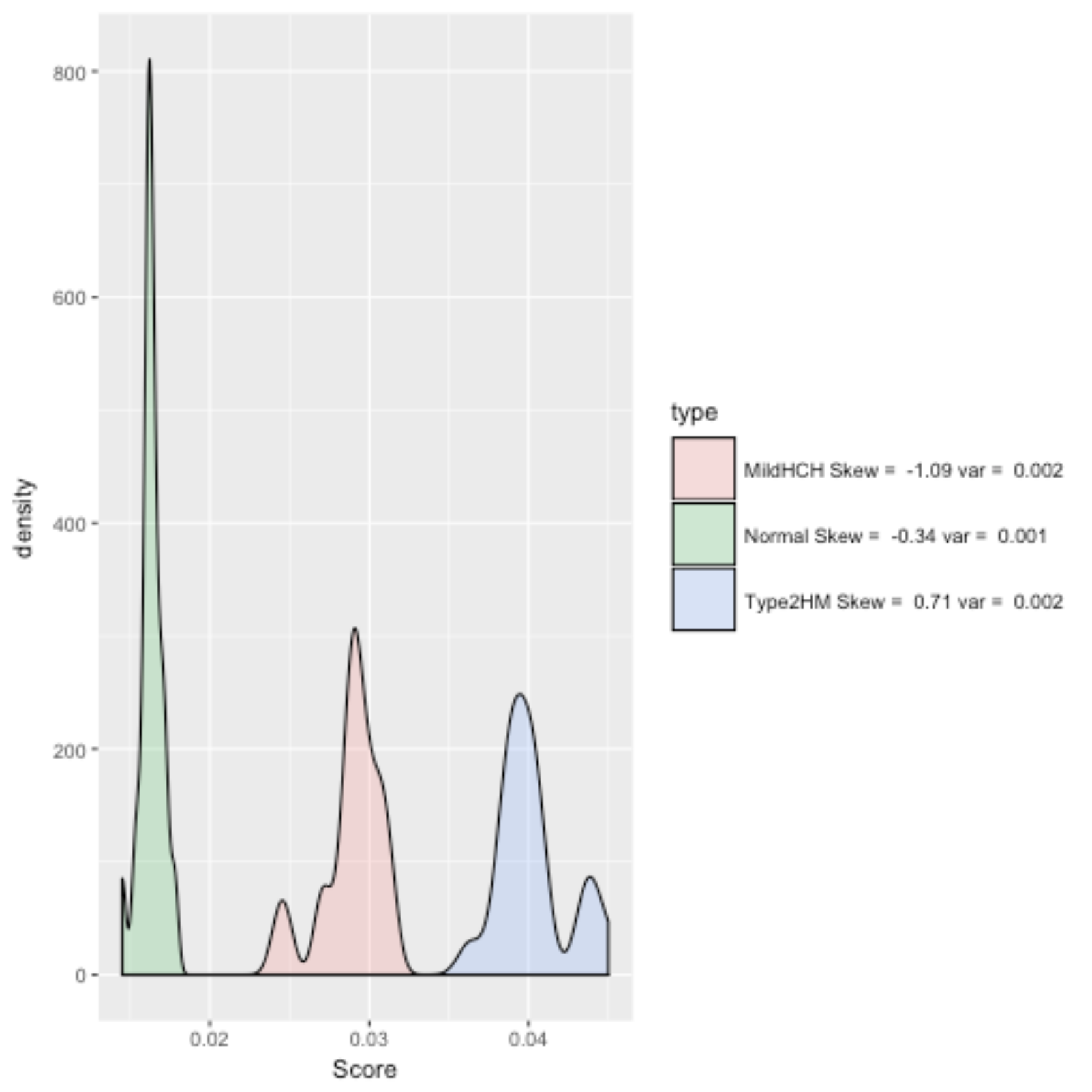}
\caption{Density Distribution of $H_{(-i)}$ }
\label{fig:HDistribution} 
\end{figure}

To resolve this we analyze the distribution of $T_{(-i)}$ in every stage of outlier elimination. For a normal subject, elimination impact of any landmark on the symmetry of the system must be more or less comparable. Hence, we should expect $T_{(-i)}$ to be symmetric about the mean with very small variance. But for a subject with mild deformity affecting only a few regions of a face, we can expect a higher variance and $T_{(-i)}$ skewed towards left, since removing landmarks that represent the affected regions will make the system strikingly symmetric as oppose to their counterparts that represent unaffected regions. We can also expect right skewed distribution of $T_{(-i)}$ when several regions of face are affected with different levels of asymmetry, see Fig \ref{fig:HDistribution}. Every time we eliminate a new outlier from the system we get a new system for which we can get new $T_{(-i)}$ scores, see Step 1. We record these scores for each stage and keep dropping landmarks till the minimum number of landmarks required to find an acceptable midsagittal plane is reached (empirical evidence as well as discussions with clinicians suggest a minimum of 8 landmarks as a good choice).  A system for which the distribution of $T_{(-i)}$ has small variance and least skewness is the best system to be used since it implies this system has most uniform landmark-elimination impact on the symmetry of the system. Hence, we pick the system that returns small variance (var $\leq$ 1.5*least variance) and least skewness for the distribution of $T_{(-i)}$.  This system is then used in step 3 to compute the midsagittal plane. 

\subsection*{ Step 3: Calculating the Midsagittal Plane}\label{sec:midsagittal}
The best choice for a midsagittal plane for a skull would be a plane for which each unpaired landmark lies on the plane, each pair of paired landmarks is equidistant from the plane and the line segments joining each pair of paired landmarks are perpendicular to the plane.  We take advantage of these heuristics to derive an optimization function, which should be minimized to yield a plane of best fit~(\ref{equ: objword}). 
\begin{multline}\label{equ: objword}
\small
 f(P) = \sum_{i=1,\dots, M} \textrm{(weighted) } Dist^2 (P,U_i) +\\  \sum_{i=1\dots N} \textrm{(weighted) } Dist^2(P,\frac{P^R_i+P^L_i}{2}) + \\  
 \gamma \sum_{i=1\dots N} \textrm{ (weighted) }Angle(n,P^{R}_i-P^{L}_i),
\end{multline} 
where  $P$ is a potential midsagittal plane, $n$ normal to $P$, $U_i$ are unpaired landmarks, $\frac{P^R_i+P^L_i}{2}$ are midpoints of the line segments joining paired landmarks and  $P^{R}_i-P^{L}_i$ is the direction of the line segments joining paired landmarks. The  $Dist$ function computes the distance between a plane and a point, the  $Angle$ function computes the angle between two vectors and  $\gamma$ is a parameter used to ensure a rough comparability of the weighting of angle and distance measurements. Weights associated with each landmark are an additional feature that can be incorporated by users if there is a bias towards certain regions of the face. For example, in certain surgeries it would be easier to modify mandible and maxilla regions as oppose to cranium regions, in which case landmarks representing cranium region could be given more weight to obtain midsagittal plane biased towards cranium region. However, for the scope of this paper we do not assign different weights to landmarks. 

Our goal is to find a plane that minimizes (\ref{equ: objword}), i.e. a plane which is a solution to the following non linear objective problem: 
\begin{equation}
\small
\begin{aligned}
&\min_{(n,d)} f =& \frac{1}{M+N}\sum_{(x) \in S_1}w_x(x^Tn+d)^2\; + \\
&&\frac{\gamma}{N}\sum_{(x) \in S_2} p_x \left (sin ^{-1} \left(  \frac{\|x \times n \|}{\|x\|} \right) \right)^2\\
& \text{subject to } 
& \|n\|^2 - 1= 0
\end{aligned}
\label{equ:objfun}
\end{equation}
where $n$ is a normal to the potential midsagittal plane, $d$ is the distance between the plane and the origin,  
$S_1 = \{U_1, \dots U_M,\frac{P_1^R+P_1^L}{2}, \dots, \frac{P_N^R+P_N^L}{2} \}$, 
$S_2 = \{ P_1^R-P_1^L, \dots, P_N^R-P_N^L \}$, parameters $w_x$ and $p_x$ are weights assigned to the landmark associated  with $x \in S_1 \textrm{ or } S_2$. Note: the elements of $S_1$ and $S_2$ are in $\mathbb{R}_3$. 

The above minimization problem (\ref{equ:objfun}) is  non-linear. Hence, we use a standard gradient descent method to obtain an optimal solution. 
There are two main issues to be taken care: first a good choice of a plane for initialization, and second the choice of $\gamma$. 

Notice that when $\gamma = 0$, the problem is quadratic and has a unique solution. Thus a unique plane, $P_0$, is obtained.  For a symmetric system this would give us the desired midsagittal plane. However, for a deformed face we can use the plane, $P_0$, as the initial midsagittal plane. 

The quantity $\gamma$ weights the effect of the discrepancy of the Euclidean distance term and the angle term in the objective function.  The numerical value of $\gamma$ depends on the scaling of the coordinate system. For units in millimeter the associated deformed model's Euclidean distances used in the objective function have an average of the order of 5 units.  In our objective function we use distance squared so such a distance will equate to a number roughly of the order of 25.  Also the angular measurements used in objective function are of the order 3 degrees. Hence to balance the two terms we chose $\gamma $ to be around $480$ ($25 \times  \frac{180}{3\pi} \approx 480$). After extensive numerical experimentation we found that the choice of $\gamma$ works well for all the models we have considered and results are robust to the choice of $\gamma$ of this order. 
 
Remark: The angle between $P_0$ and the plane which is the solution to the optimization problem in (\ref{equ:objfun}) is small. If we restrict our search for  planarsolutions close to  the plane $P_0$,  we may use the approximation $\sin (\theta) \approx \theta$. This approximates the original problem (\ref{equ:objfun}) by  a quadratic problem.
\begin{thm}  Existence and Uniqueness  theorem \label{Thm: Existence}\end{thm} 
If we consider the quadratic system given by the approximation  $sin(\theta) \approx \theta$ then (\ref{equ:objfun}) will have a unique solution up to the sign of the normal vector if the landmarks are ``generic'' (put generic in quotes). Our notion of genericity is that the geometric multiplicity of the smallest eigenvalue of a certain matrix derived from the landmark positions is 1. (see \nameref{App_1})
\section*{Validation}
\begin{figure}[!t]
\centering
\subfloat[Symmetric skull]{\includegraphics[width=1.4in,trim={4.5cm 3cm 4.5cm 3cm},clip]{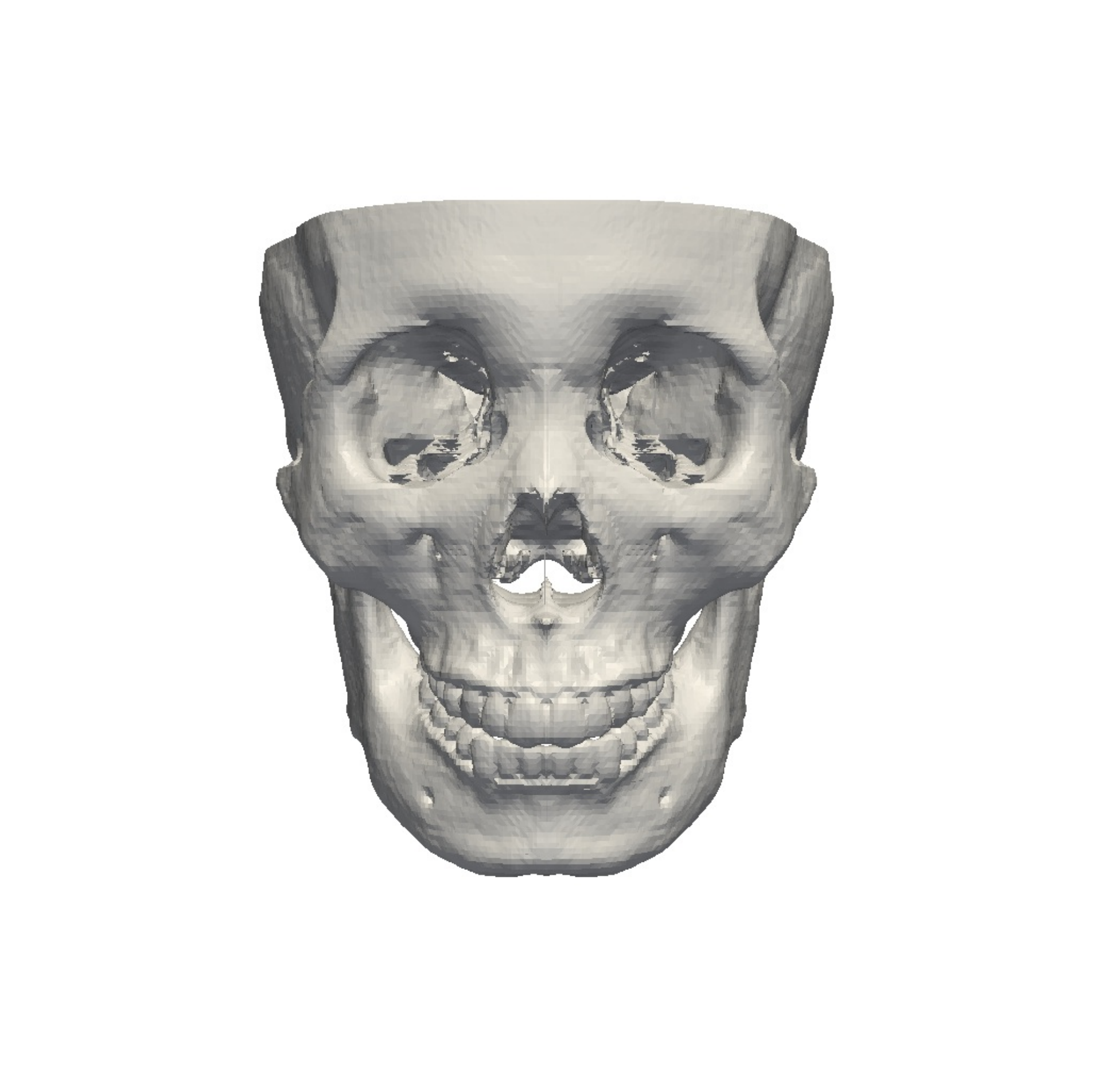}
\label{fig:Shangha002Sym}}
\hfill 
\subfloat[Horizontal Condylar  Hyperplasia]{\includegraphics[width=1.4in,trim={4.5cm 3cm 4.5cm 3cm},clip]{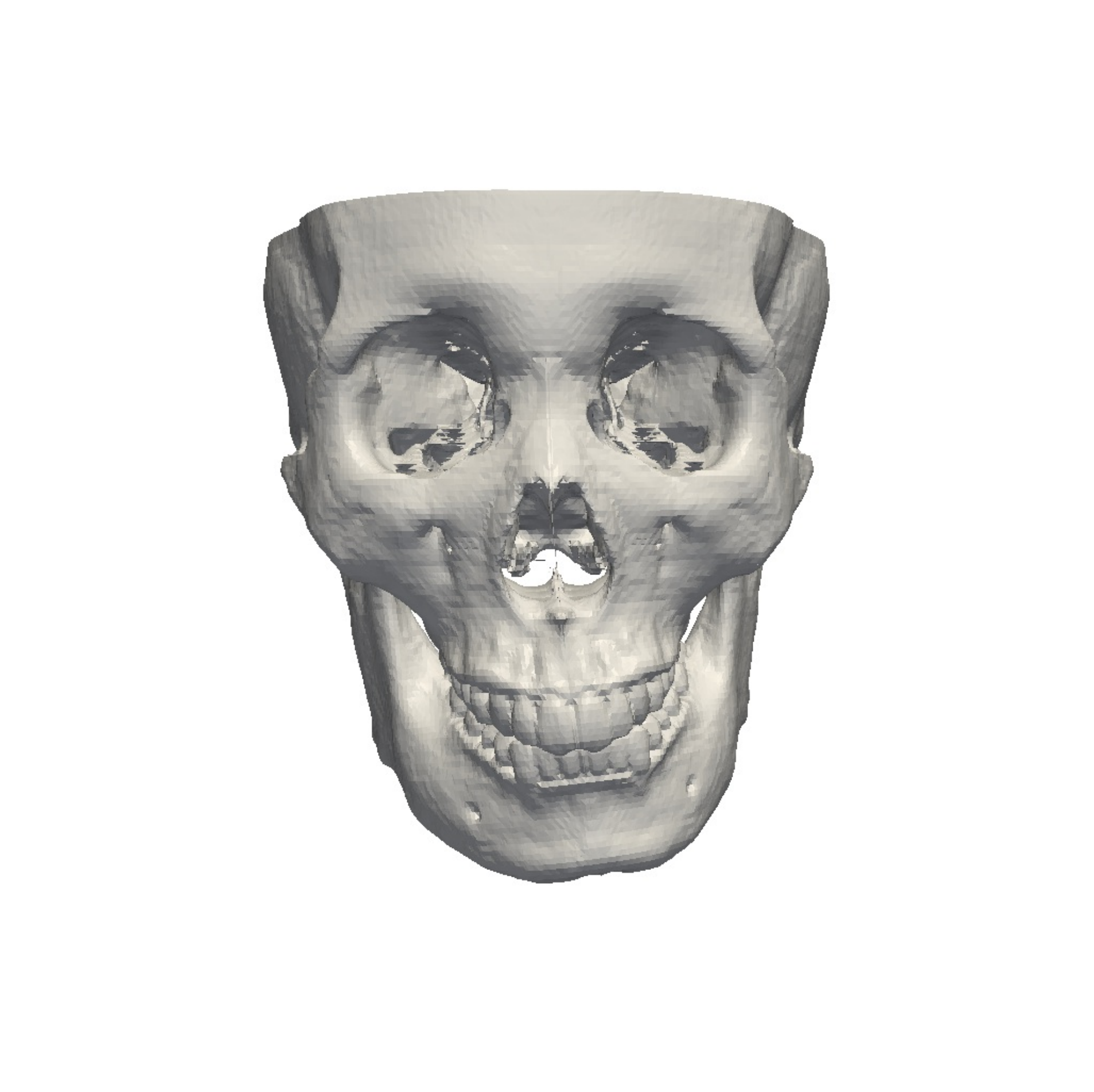}
\label{fig:Shanghai002HoriCond}}
\hfill 
\subfloat[ Vertical Condylar Hyperplasia]{\includegraphics[width=1.4in,trim={4.5cm 3cm 4.5cm 3cm},clip]{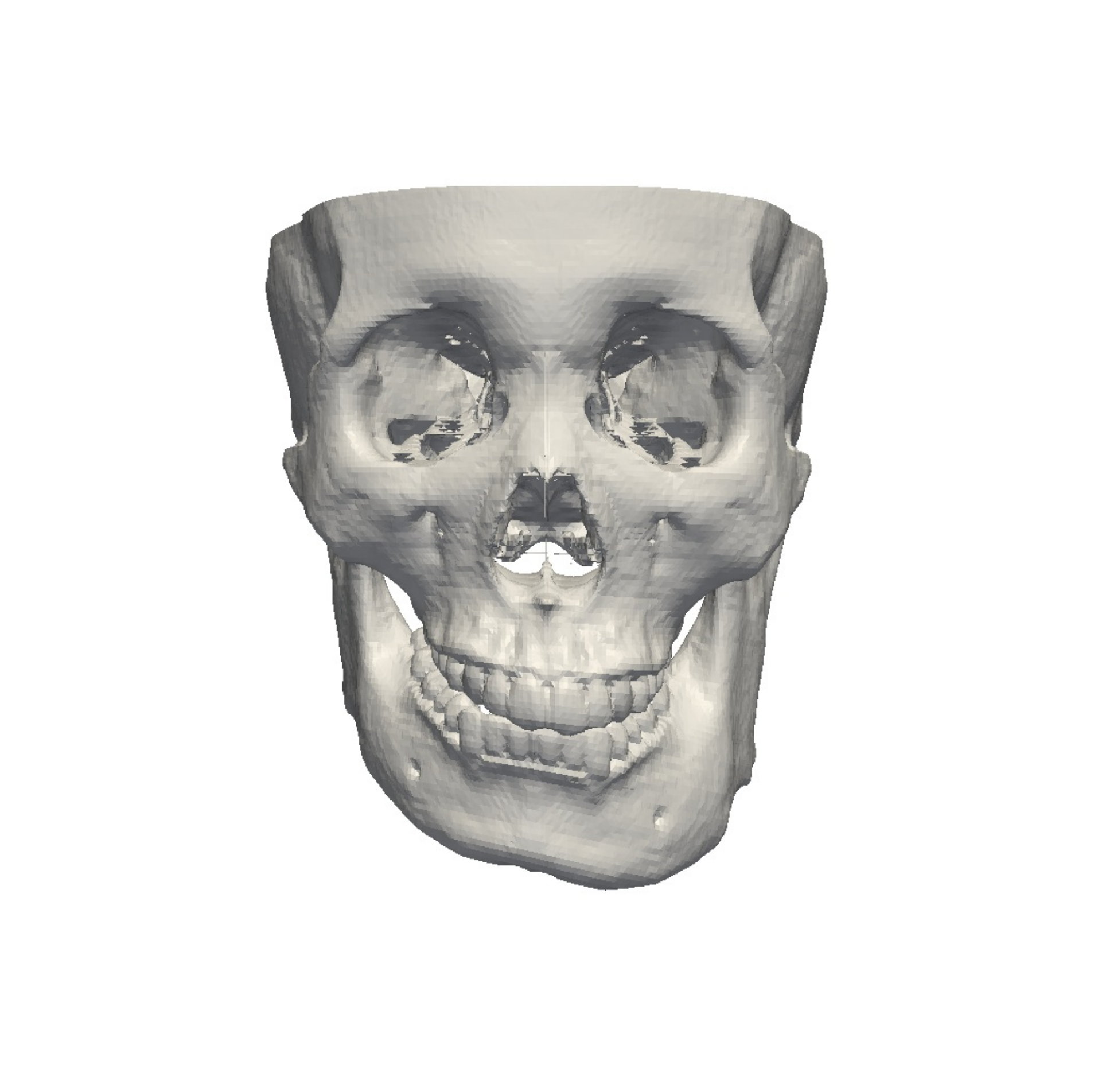}
\label{fig:Shanghai002VerCond}}
\hfill
\subfloat[Type1 Hemi Micro]{\includegraphics[width=1.4in,trim={4.5cm 3cm 4.5cm 3cm},clip]{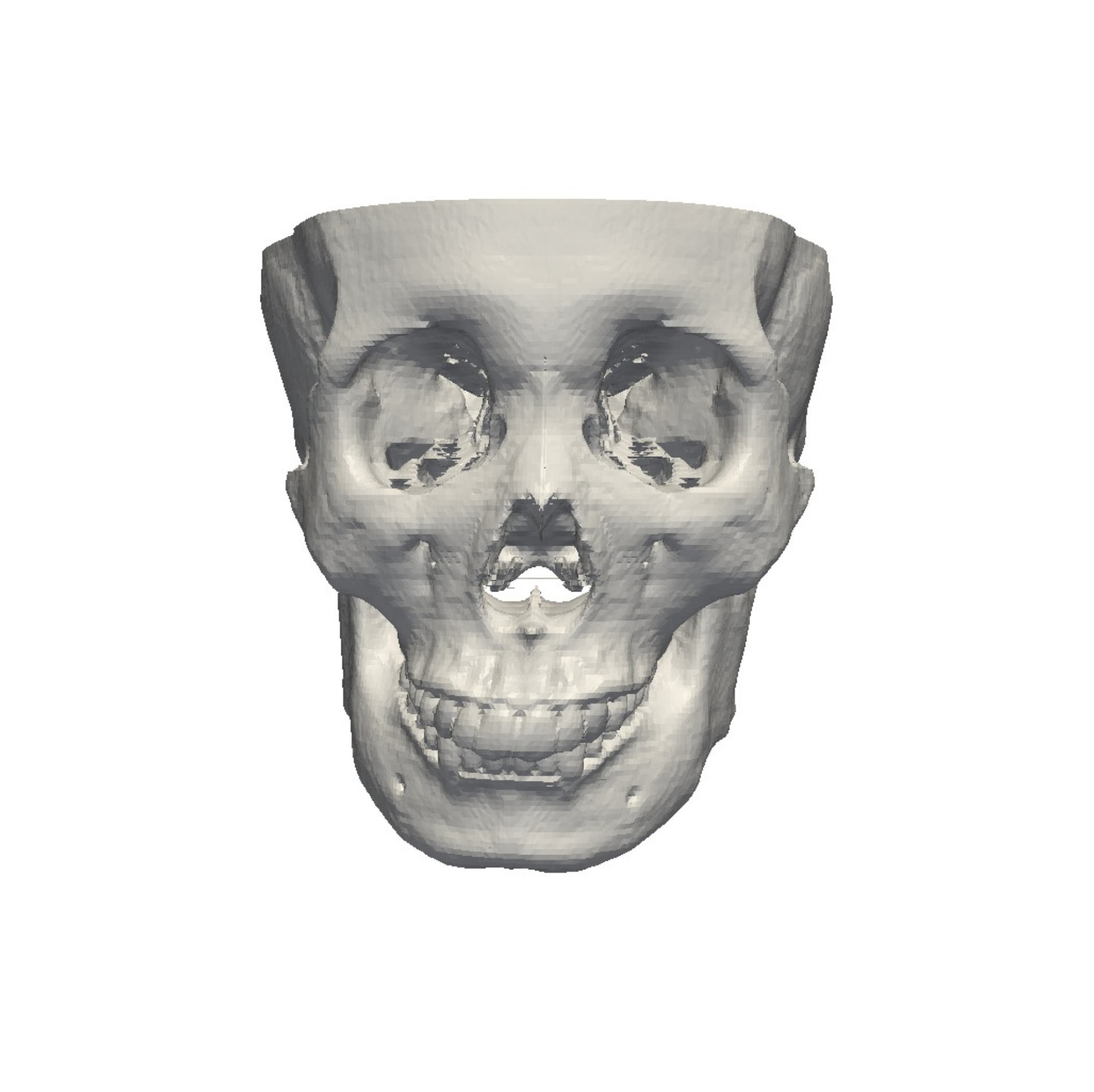}
\label{fig:Shanghai002Type1}}
\hfill
\subfloat[Type2 Hemi Micro]{\includegraphics[width=1.4in,trim={4.5cm 3cm 4.5cm 3cm},clip]{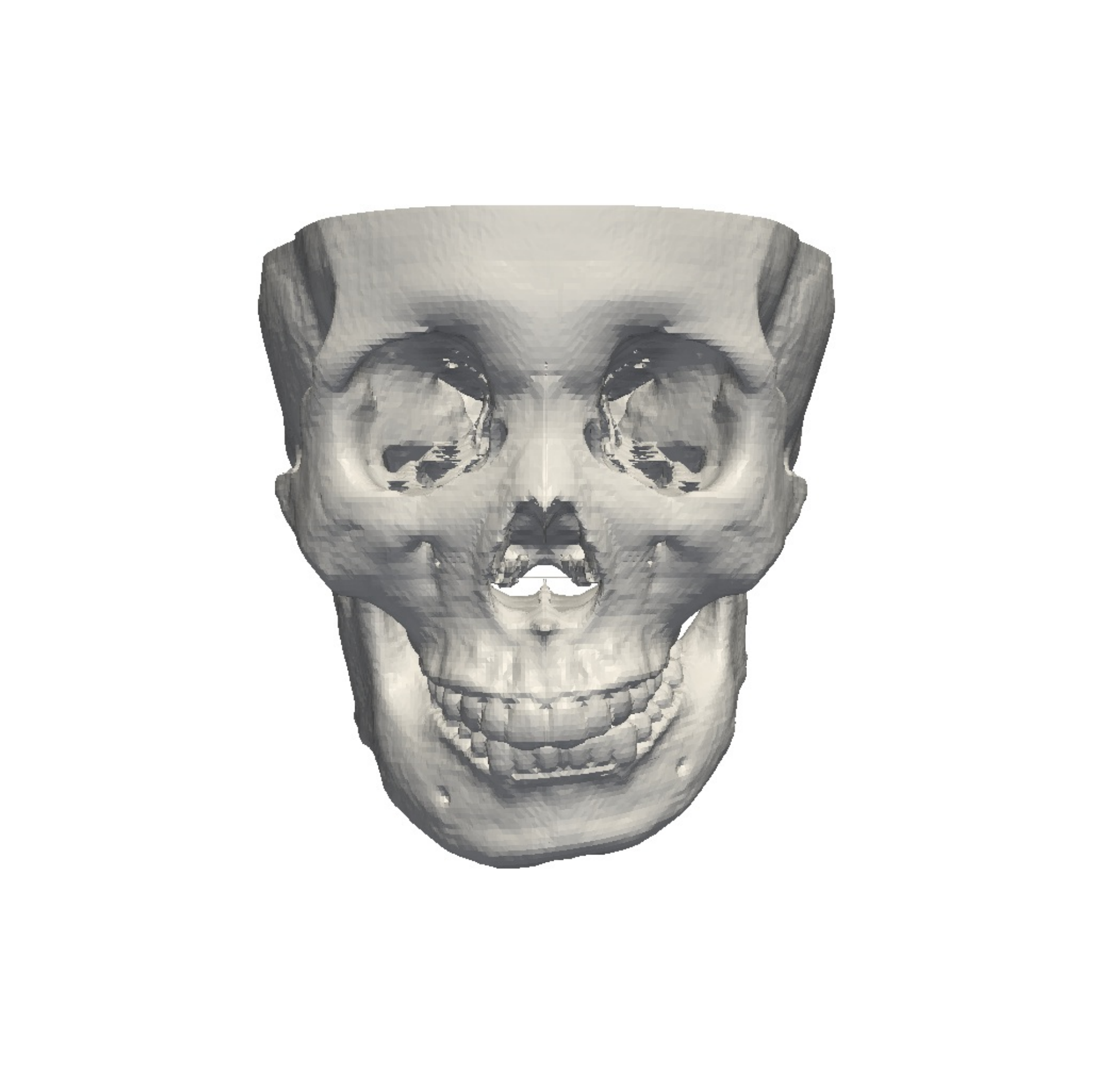}
\label{fig:Shanghai002Type2}} 
\caption{A symmetric skull and its manually morphed versions resembling 4 deformities:  The left most skull is a perfectly symmetric skull, generated by replacing the  left side of the normal subject~\ref{fig:normalcase} by the mirror image of the right side.  The next 4 skulls are manually deformed skulls mimicking 4 deformities starting from left, horizontal condylar hyperplasia, vertical condylar hyperplasia, type1 hemifacial microsomia and type 2 hemifacial microsomia. }
\label{fig:SymAnd4Deformities}
\end{figure}

Our algorithm was based on the assumption that the optimal midsagittal plane of a patient with a deformity is a hypothetical premorbid midsagittal plane of the patient. To this end, we developed a novel method to test the efficacy of our algorithm. In this approach, a perfectly symmetric skull model without fluctuating asymmetry, was first generated using a normal subject Fig~\ref{fig:CTtoBone}. The midsagittal plane of this symmetric model serves as the ground truth. In the next step, the perfectly symmetric model was morphed to emulate different facial asymmetrical deformities Fig~\ref{fig:SymAnd4Deformities}. The midsagittal plane for these models was then estimated using LAGER. Finally, the algorithm-generated midsagittal plane was compared with the ground truth.  

The algorithm was validated on 4 types of deformities:  hemifacial microsomia (Type1 and Type 2), and unilateral condylar hyperplasia (horizontal and vertical), Fig~\ref{fig:SymAnd4Deformities}. 
Five 3D CT skull models of normal subjects were randomly selected from a large CT data archive. 3 oral surgeons and orthodontists (for a different project) evaluated these normal subjects. Each subject had a normal facial appearance, no congenital or acquired CMF deformity, no history of facial surgery or trauma, and had type 1 (normal) dental occlusion. The CT scan was acquired using a spiral CT scanner with 1.25mm of scanning thickness. The scanning matrix was $512\times 512$. The subject was in supine position with the upper and the lower teeth biting in maximal intercuspation during the scanning. The data collection was approved by Ethic Committee of Shanghai Ninth People's Hospital. The CT data was HIPAA de-identified prior to the project. 
In order to create a perfectly symmetric skull, we split the 3D skull model of a normal subject along an initial midsagittal plane that divides the skull into best possible identical right and left halves. The left side of the skull was then replaced by the mirror image of the right side of the skull to obtain a perfectly symmetric skull. Two oral surgeons (K.C. and J.X.) digitized the landmarks. The unpaired landmarks were perfectly located on the midsagittal plane while the paired landmarks located on the right and left side of the skull as if they were mirror imaged. The landmarks were then digitally  ``glued" onto the model. These midsagittal planes and the landmarks served as ground truth. 

This symmetric model, together with the digitally ``glued" landmarks,  was then deformed manually by two oral surgeons (K.C. and J.X.) using free form deformation techniques (3D Studio Max, Autodesk, Inc., San Rafael, CA), to generate a skull mimicking a given deformity as mentioned above, based on clinical definitions~\cite{MCC90,NKB08}, Fig(\ref{fig:SymAnd4Deformities}). Therefore, for each subject, there was one perfectly symmetric model and 4 deformed models representing the 4 deformities. A third oral surgeon (J.G.) verified these morphed models to ensure that these models mimicked a real patient's deformities. During the morphing, the landmarks were also moved to new locations along with the model. The coordinates of these landmarks represented the skulls with deformities, which served as an experimental group. We added Gaussian noise (mean 0mm and standard deviaiton 3mm) to the regions which were not morphed to model the presence of fluctuating asymmetry in real patients. We use LAGER algorithm to generate midsagittal plane. For the computations we simply assign $0$ weight for the landmarks detected as outliers and 1 for the rest of landmarks. 

 In order to assess whether the algorithm-generated midsagittal plane of these models could be used clinically, four outcome variables, one angular and three linear measurements were used. They are described as follows:
\begin{subequations}
\begin{align}
\theta &= \textrm{Angle between } G \textrm { and } MP\;\; \\ DistPg &= \textrm{Distance between Pg and } MP \\ \;\;\;  DistU1 &= \textrm{Distance between U1 and } MP \\
DistN &= \textrm{Distance between N and } MP
\end{align}\label{equ: perCri}
\end{subequations}\\
where $MP$ is the algorithm-generated midsagittal plane for a given deformed model; $G$ is the ground truth midsagittal plane for the corresponding symmetric model;  $N,\;U1$ and $Pg$ are the positions of  the naison, the central dental midline and the pogonion on the corresponding symmetric model. Based on clinical experience we set the following criteria for a plane to be clinically acceptable as a midsagittal plane.  
\begin{equation}
\footnotesize
\theta < 2^ \circ, \; DistN < 1\textrm{mm}, \; DistU1 < 1\textrm{mm},\; DistPg < 2\textrm{mm}
\label{equ:perCriInq}
\end{equation}
Tables (\ref{tb:HorCondHype}-\ref{tb:Type2HemiMicro}) presents the 4 outcome variables as described in (\ref{equ: perCri}). 
These tables show the quality performance of the algorithm on 5 artificial models for each of the 4 deformities. We can see that all the LAGER algorithm-generated midsagittal planes satisfy the criteria to be qualified as clinically acceptable. 
\begin{table}[!t]
\renewcommand{\arraystretch}{1.3}
\caption{Horizontal Condylar Hyperplasia} \label{tb:HorCondHype}
\centering
\begin{tabular}{c|c|c|c|c}
\hline
& $\theta $ & DistN& DistUone & DistPg \\
\hline
Case 1& 0.13 &  0.18 &   0.36 & 0.4 \\
Case 2&    0.15   & 0.43   & .44    &.5 \\
    Case 3 & 0.26 &  0.24 &  0.61   &0.7 \\
Case 4&    0.20  &  0.10  &  0.40  &  0.5 \\
Case 5 &    0.20&   0.01 &  0.25  &0.3 \\
    \hline
    \end{tabular}
    \end{table}
\begin{table}[!t]
\renewcommand{\arraystretch}{1.3}
\caption{Vertical Condylar Hyperplasia} \label{tb:VerCondHype}
\centering
\begin{tabular}{c|c|c|c|c}
\hline
& $\theta$ & DistN & DistUone & DistPg \\
\hline
Case 1& 0.10 &  0.05 &   0.08 & 0.12 \\
Case 2&    0.17   & 0.06   & 0.17    &0.23 \\
    Case 3 & 0.03 &  0.04 &  0.01   &0.01 \\
Case 4&    0.15  &  0.02  &  0.2  &  0.26 \\
Case 5 &    0.06&   0.01 &  0.06  &0.09 \\
    \hline
    \end{tabular}
    \end{table}
\begin{table}[!t]
\renewcommand{\arraystretch}{1.3}
\caption{Type 1 Hemifacial Microsomia} \label{tb:Type1HemiMicro}
\centering
\begin{tabular}{c|c|c|c|c}
\hline
& $\theta$ & DistN & DistUone & DistPg \\
\hline
Case 1 & 0.41 &  0.31  &  0.22 & 0.42 \\
Case 2 & 0.15  & 0.44  & 0.24    &0.18 \\
Case 3 & 0.11 &  0.27  &  0.38   &0.41 \\
Case 4 & 0.07  &  0.04 &  .06  &  .09 \\
Case 5 & 0.06 &   0.07 &  0.76   &1 \\
    \hline
    \end{tabular}
    \end{table}

\begin{table}[!t]
\renewcommand{\arraystretch}{1.3}
\caption{Type 2 Hemifacial Microsomia} \label{tb:Type2HemiMicro}
\centering
\begin{tabular}{c|c|c|c|c}
\hline
& $\theta$ & DistN & DistUone & DistPg \\
\hline
Case 1& 0.15 &  0.43 &   0.64 & 0.71 \\
Case 2&    0.72   & 0.66   & 0.33    &0.59 \\
    Case 3 & 0.06 &  0.80 &  0.84   &0.84 \\
Case 4&    0.22 &  0.37 &  0.07  &  0.02 \\
Case 5 &    0.58&   0.67 &  0.11  &0.35 \\
    \hline
    \end{tabular}
    \end{table}
\section*{Discussion}
The problem of estimating the best plane of symmetry arises in various areas such as morphometrics~\cite{KLI98,KBM02}, biomedical applications~\cite{TCB,WTH11} and surgical planning~\cite{XMG11}. Although a number of different approaches have been studied, in the application of planning CMF surgeries, we found two approaches to be the most relevant: the first approach is based on Procrustes superimposition~\cite{KLI98}  and the second approach is based on finding  the  optimal plane by minimizing various combinations of Euclidean distances and  angles between the potential plane and  data points. 

The Procrustes method has been used extensively in morphometric shape analysis (for example see \cite{KLI98}).  The advantage of this method is that it provides a preliminary plane that is close to the midsagittal plane we are interested in. However, if there are outliers among the set of landmarks, they may greatly affect the choice of plane (sometimes called the Pinocchio effect ~\cite{ZSS04}). But, a recursive Procrustes method may be used to rank the landmarks and detect outliers as explained in~\cite{GJN15}.
Also, weighted Procrustes methods are found in literature that could potentially help but assigning weights in our context is not obvious. Similarly, parallels to identifying outliers in Step 2 of LAGER are not obvious for Procrustes method. 

P. Claes et. al~\cite{CWV11} have used another very interesting approach to resolve the shortcomings of  Procrustes method. Instead of using a coarse mesh of anatomical landmark points, they use a dense mesh of about 10,000 points approximating the surface of a face, which they call "quasi-landmarks". Obtaining quasi-landmarks with manual tagging isn't effective. To obtain a set of quasi-landmarks for a face, they map a known set of quasi-landmarks of an average normal face onto the patients face using registration techniques. They have reported that different registration algorithms yield different results and also the correspondence between the points is basically a loose approximation. Besides this is a computationally expensive procedure. In the next step they use a robust Procrustes method. This is an iterative process; in every iterate they divide the quasi-landmarks into inliers and outliers, and assign weights to landmarks based on the distribution of the matching quality measure for each landmark. They stop the iteration when the set of inliers and the set of outliers stabilize and then use a weighted Procrustes method to obtain the midsagittal plane. The iterative process to identify outliers and inliers can be compared to the first two steps of LAGER algorithm; however, their statistical analysis requires large number of landmarks to obtain stable parameter estimations and hence it is not directly applicable for our situation. 

\begin{figure}[!t]
\centering
\subfloat[Outliers were dropped]{\includegraphics[width=1.5in,trim={4.5cm 3cm 4.5cm 3cm},clip]{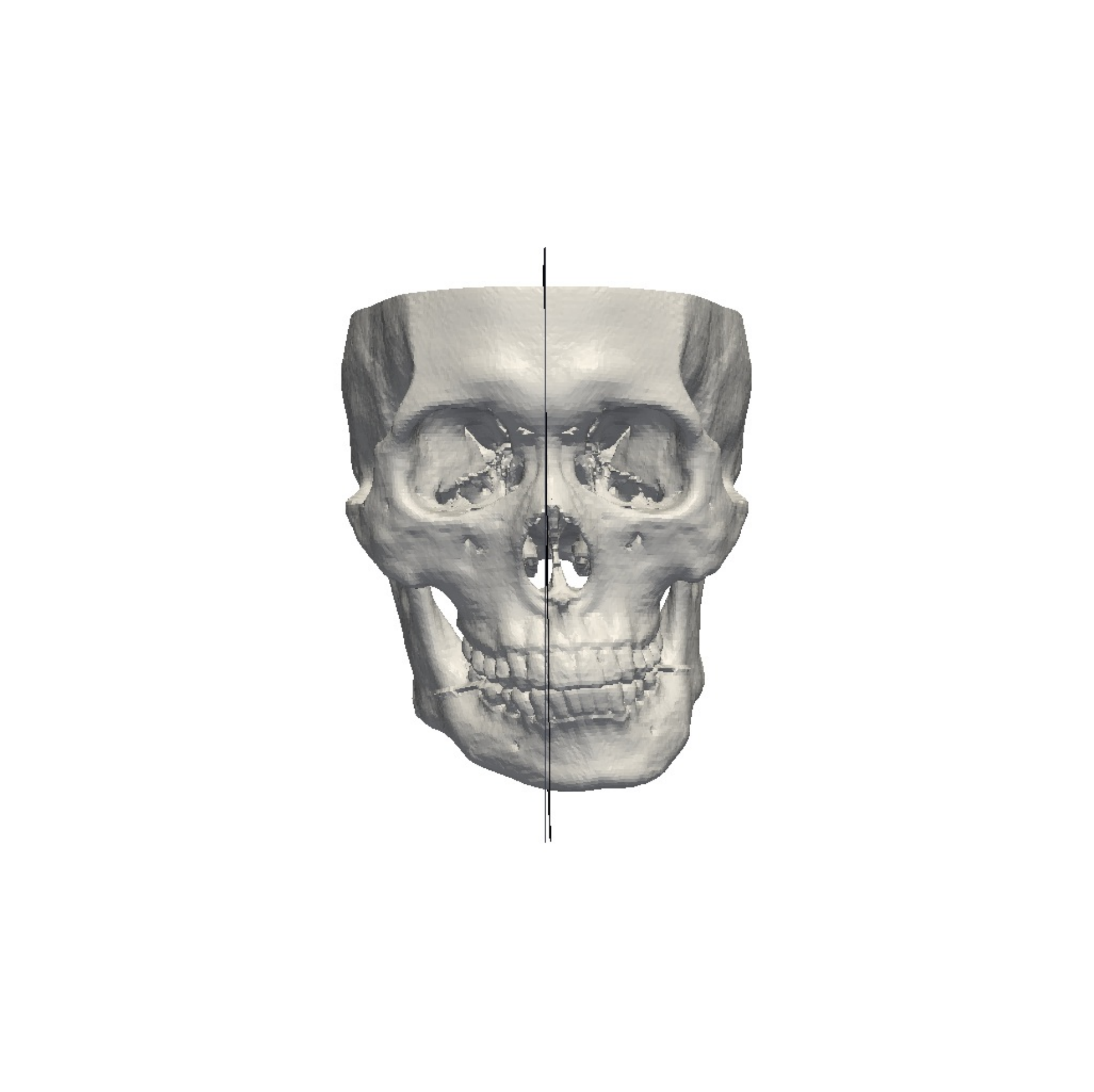}%
\label{fig:DefoModelSkullFront}}
\hfil
\subfloat[Outliers were not dropped]{\includegraphics[width=1.5in,trim={4.5cm 3cm 4.5cm 3cm},clip]{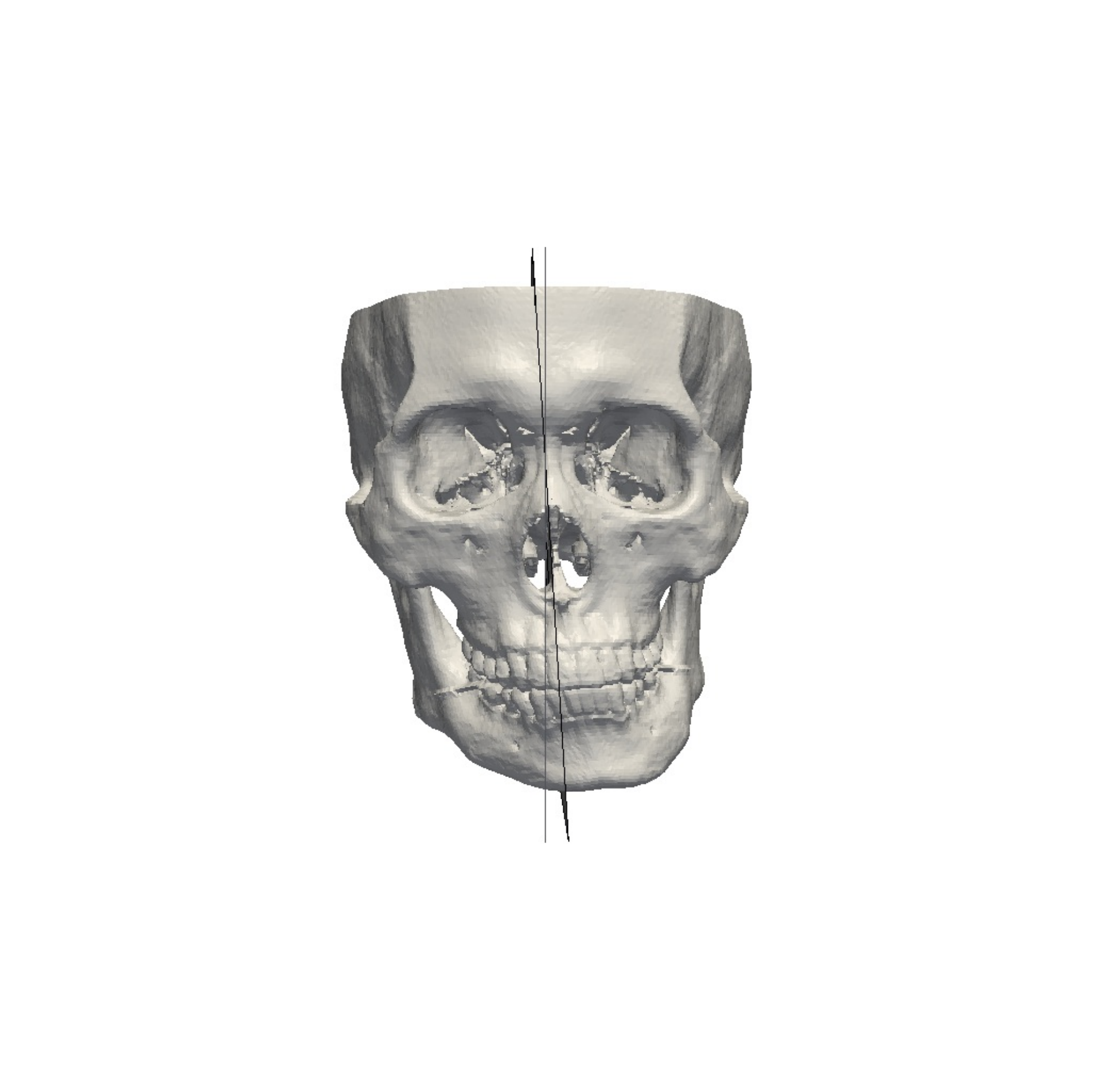}%
\label{fig:DefoModelTwoPlaneNoDrop}}
\caption{Deformed skull, algorithm-generated plane (grey), ground truth plane (black): The skull resembles a type 2 hemifacial microsomia. The planes in black are the algorithm-generated planes and the plane in grey is the ground truth.  To the left we have algorithm-generated plane, clearly the two planes are indistinguishable. To the right,  the plane in black is calculated by our algorithm without dropping outliers. One can clearly see that the plane is not optimal.}
\label{fig:DefoModelTwoPlanes}
\end{figure}

A second approach similar to what that we adopted, comprises of minimizing a combination of terms involving angles between line segments joining paired points and the proposed plane, and the Euclidean distances between the points and the proposed plane  (for other representative studies see~\cite{WTH11,CHW08}).  However, these earlier studies did not consider dropping outliers according to their level of asymmetry, which greatly improves the midsagittal plane calculation, as shown in Fig~\ref{fig:DefoModelTwoPlanes}.
We developed the LAGER algorithm based on the second approach by incorporating ranking landmarks and dropping outliers in the algorithm. This results in a more accurate midsagittal plane. As shown in Fig~\ref{fig:DefoModelTwoPlanes}, dropping the outliers yields better results. In the future we will further validate our LAGER algorithm using a larger group of human subjects. One of the limitations of LAGER method is that it can not identify midsagittal plane when deformity is sever and affects all the regions. But LAGER is capable of identifying such cases and gives a warning to user.  We intend to use machine learning techniques on large experimental data sets to obtain a good choice of weights for landmarks that could improve the results in such extreme cases. 

Accurate digitization of landmark is essential for assessing CMF anatomy and plan surgeries. J. Xia ~\cite{ZGT15} have proposed effective methods to allow automatic digitization of landmark in clinical use. Hopefully in near future these methods could be combined with our algorithm to achieve a completely automated algorithm to obtain midsagittal plane. 
\section*{Conclusion}
We have developed a landmark-based algorithm to automatically calculate the midsagittal plane of human skulls affected by deformities. We validated the algorithms on 4 different types of CMF deformities using synthetic human skulls. The algorithm detects correctly the least asymmetric regions of the face, which are then used to compute midsagittal plane. All the computed midsagittal planes satisfied the clinical criterion to be qualified as clinically acceptable. Thus it can be used clinically to determine the midsagittal plane for patients with CMF deformities. 

\section*{Supporting Information}
\paragraph{Appendix: Proof of  theorem \ref{Thm: Existence}}
\label{App_1}
\begin{equation}
\begin{aligned}
&\min_{n,d} f = &\frac{1}{M+N}\sum_{x \in S_1}w_x(x^Tn+d)^2\; + \\
&&\frac{\gamma}{N}\sum_{(y) \in S_2} p_y  \left( \frac{(y \times n)^T(y \times n)}{\|y\|} \right) \\
&\text{subject to } \\
&&\|n\|^2 - 1= 0
\end{aligned}\label{equ:objfunLin}
\end{equation}
We use Lagrange multipliers  to obtain first order optimality conditions along the lines of  the   proof given in~\cite{CHW08}. The Lagrangian function associated to the above system is
\begin{equation}
\begin{split}
\mathcal{L}(n,d,\lambda) = \frac{1}{M+N}\sum_{x \in S_1}w_x(x^Tn+d)^2\; + \\
\frac{\gamma}{N}\sum_{y \in S_2} p_y  \left( \frac{n^T [y]^T_{\times} [y]_{\times} n)}{\|y\|^2} \right) + \lambda(1-\|n\|^2)
\end{split}
\label{equ:Lagarangian}
\end{equation}
Here $[y]_{\times}$ is the cross product matrix of $y$, i.e. $a \times y = a [y]_{\times}$, for any vector $a$. 
Setting $\frac{\partial \mathcal{L}}{\partial d} = 0$, we get

\begin{equation}
 d = -\dfrac{\sum \limits_{x \in X} w_x x^Tn }{\sum \limits_{x\in S_1} w_x}
 \label{equ:FirstOrderOptd}
\end{equation}
Replacing the expression for $d$ from (\ref{equ:FirstOrderOptd}) we obtain
\begin{equation}
\begin{split}
\mathcal{L}(n,\lambda) = \frac{1}{M+N}\sum_{x \in S_1}w_x \left (x^Tn -\dfrac{\sum \limits_{z \in S_1} z^Tn }{\sum \limits_{v \in S_1} w_v} \right )^2 \; + \\
\frac{\gamma}{N}\sum_{y \in S_2} p_y  \left( \frac{n^T [y]^T_{\times} [y]_{\times} n)}{\|y\|^2} \right) + \lambda(1-\|n\|^2)
\end{split}
\end{equation}
We can rewrite the above expression as 
\begin{equation*}
\footnotesize
\begin{split}
\mathcal{L}(n,\lambda) = \hspace{200pt}\\ n^T \frac{1}{M+N} \left (\sum \limits_{x \in S_1}w_x x x^T-\dfrac{ \sum \limits_{x \in S_1}\sum \limits_{z \in S_1}w_xw_z xz^T }{\sum \limits_{v \in S_1}w_v} \right )n \; + \\
n^T \left (\frac{\gamma}{N}\sum_{y \in S_2} p_y  \left( \frac{ [y]^T_{\times} [y]_{\times}}{\|y\|^2} \right)\right)n + \lambda(1-\|n\|^2).
\end{split}
\end{equation*}
Hence, the derivative of the Lagrangian with respect to $n$ is
\begin{equation}
\dfrac{\partial \mathcal L}{\partial n} = (B-\lambda I)n,
\end{equation}
where \\ \footnotesize $ B = \frac{1}{M+N} \left (\sum \limits_{x \in S_1}w_x x x^T-\dfrac{1}{\sum \limits_{v \in S_1}w_v} \sum \limits_{x \in S_1}\sum \limits_{z \in S_1} w_xw_zxz^T  \right )+  \frac{\gamma}{N}\sum \limits_{y \in S_2} p_y  \left( \frac{ [y]^T_{\times} [y]_{\times}}{\|y\|^2} \right).$
\normalsize
Hence, the Lagrange multipliers must be eigenvalues of the matrix $B$. Upon evaluating the Lagrangian we see its 
corresponding values are also the eigenvalues of $B$. So the function is optimized at the smallest eigenvalue of $B$. If $B$ is positive definite and its smallest eigenvalue's geometric multiplicity is 1 then we have a unique solution up to the sign of the normal vector.

\section*{Acknowledgments}
This study was partially supported by NIH/NIDCR R01 DE022676  Dr. Chen was sponsored by Taiwan Ministry of Education while he was working at the Surgical Planning Laboratory, Department of Oral and Maxillofacial Surgery, Houston Methodist Research Institute, Houston, TX, USA. Dr. Tang was sponsored by China Scholarship Council while he was working at the Surgical Planning Laboratory, Department of Oral and Maxillofacial Surgery, Houston Methodist Research Institute, Houston, TX, USA. 
The authors would also like to thank to Dr. Robert Azencott for several useful discussions.

%
%
%

\end{document}